# Single Channel EEG Based Insomnia Identification Without Sleep Stage Annotations

Chan-Yun Yang[1], Nilantha Premakumara[1], Hooman Samani[2] , Chinthaka Premachandra[3]

[1] *Department of Electrical Engineering, National Taipei University, New Taipei City, Taiwan*
[2] *Creative Computing Institute, University of the Arts London, London, UK*

[3] *Department of Electronic Engineering, Shibaura Institute of Technology, Tokyo, Japan*

Correspondence:
Nilantha Premakumara, Department of Electrical Engineering, National Taipei University, Email: nilaprem108@gmail.com
Hooman Samani, Creative Computing Institute, University of the Arts London, Email: h.samani@arts.ac.uk
Both the authors contributed equally in correspondence

## Abstract

This paper proposes a new approach to identifying patients with insomnia using a single EEG channel, without the need for sleep stage annotation. Data preprocessing, feature extraction, feature selection, and classification techniques are used to automatically detect insomnia based on features extracted from spectral and temporal domains, including relative power in the $\delta$, $\sigma$, $\beta$ and $\gamma$ bands, total power, absolute slow wave power, power ratios, mean, zero crossing rate, mobility, and complexity. A Pearson correlation coefficient, t-test, p-value, and two rules are used to select the optimal set of features for accurately classifying insomnia patients and rejecting negatively affecting features. Classification schemes including a general artificial neural network, convolutional neural network, and support vector machine are applied to the optimal feature set to distinguish between insomnia patients and healthy subjects. The performance of the model is validated using 50 insomnia patients and 50 healthy subjects, with the Fp2 channel and 1D-CNN classifier achieving the highest accuracy and Cohen's kappa coefficient at 97.85% and 94.15%, respectively. The developed model has the potential to simplify current sleep monitoring systems and enable in-home ambulatory monitoring.

## 1. Introduction

According to the Diagnostic and Statistical Manual of Mental Disorders V (DSM-5), a person with insomnia has trouble falling asleep, staying asleep, or both for at least three nights a week for at least three months. Insomnia can increase the risk of developing diabetes, hypertension, cardiovascular disease, and depression. It may also increase the risk of accidents and falls among adult people.

A diagnosis of insomnia is typically based on extensive subjective and objective data collected from individuals who have reported experiencing the condition. This diagnosis may include daily sleep diaries, Insomnia Severity Index (ISI) questionnaires for subjective information, and polysomnography (PSG) for objective results. However, the use of PSG for sleep monitoring is time-consuming and costly for both patients and medical professionals. While algorithmic sleep classification techniques have been developed in recent years for sleep staging and event marking, there is currently no integrated method for directly identifying insomnia. Monitoring one or two channels of an EEG may be a promising approach for identifying insomnia at home. However, only some studies have examined the use of fewer channels of an EEG for automatic insomnia identification. This study aims to develop a simple model for automatic insomnia identification using one or two channels of an EEG. By using actual clinical data collected from medical sites, the study employs a straightforward framework for identification modeling. The direct approach of this development avoids the need for sleep stage annotation, and yields improved efficacy.

Rapid Eye Movement (REM) sleep and Non-Rapid Eye Movement (NREM) sleep are two distinct phases of human sleep that eye movement patterns can distinguish. In 1968, Rechtschaffen and Kales (R & K) created a framework for classifying different stages of human sleep based on brain waves and physiological characteristics (Hori *et al.*, 2001). According to R & K, there are four stages of NREM sleep (Stages 1-4) and two types of REM sleep (phasic and tonic). The American Academy of Sleep Medicine (AASM) has made slight modifications to these standards, which are now outlined in the Manual for Scoring Sleep Stages and Associated Events (Iber *et al.*, 2007). One notable difference between the AASM and R & K standards is the combination of Sleep Stages 3 and 4. Electroencephalography (EEG) signals can be used to differentiate between sleep stages, as shown in Figure 1.

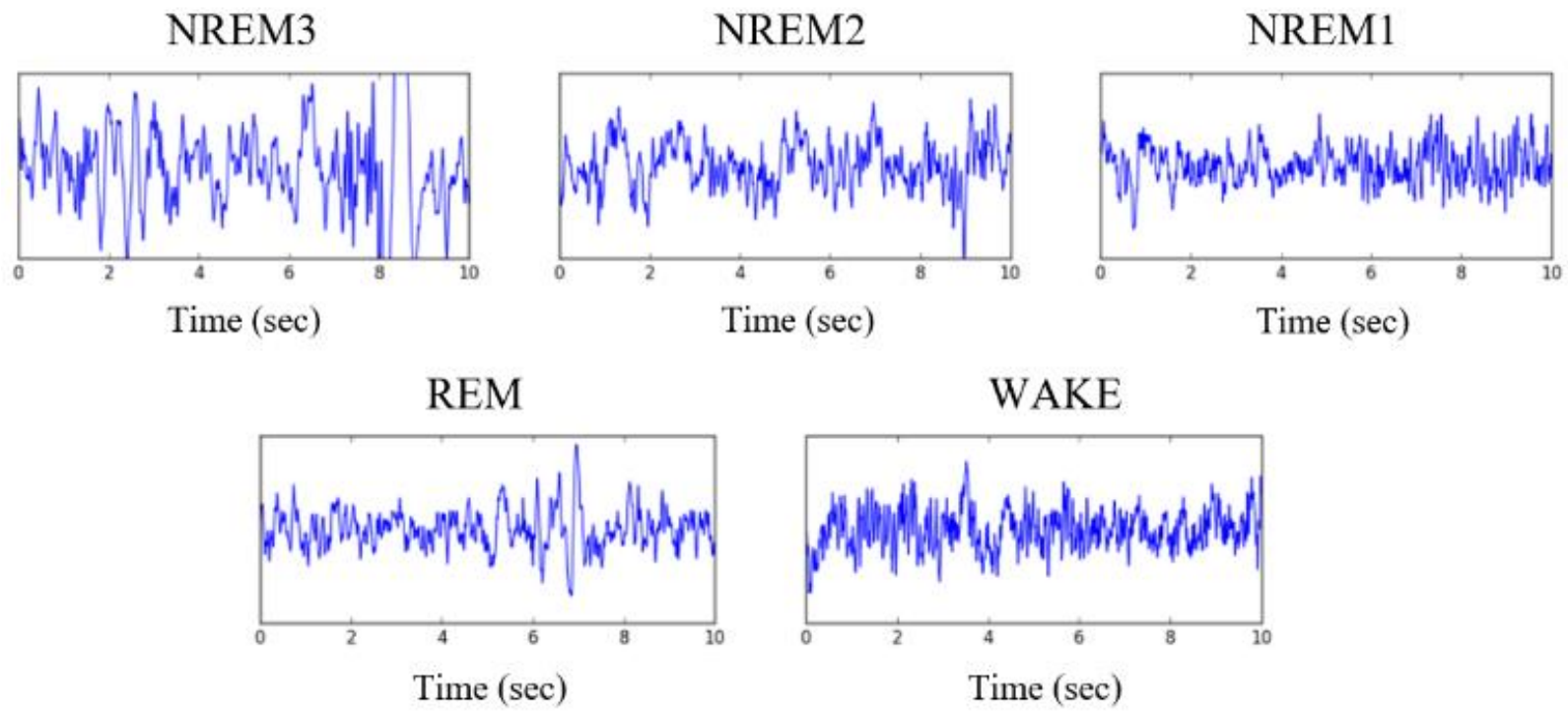

Figure 1  Rescaled time domain traces for each AASM sleep stage.

There are three classification systems used to categorize sleep disorders formally: DSM (Reynolds 3rd and O'Hara, 2013) from the American Psychiatric Association, the International Classification of Sleep Disorders (ICSD) (Mansukhani *et al*., 2014) from the American Academy of Sleep Medicine, and the International Classification of Diseases (ICD) (Edinger *et al.*, 2019) from the World Health Organization. The ICSD divides sleep disorders into six categories: insomnia, sleep-related breathing disorders, central disorders of hypersomnolence, circadian rhythm sleep-wake disorders, parasomnias, and sleep-related movement disorders. These categories are further sub-classified into approximately 60 subcategories.

Insomnia, also known as sleeplessness, is characterized by difficulty falling asleep or staying asleep for the necessary amount of time. Symptoms of insomnia may include frequent awakenings during the night, early morning awakenings, a general sense of fatigue, poor attention skills, and a feeling of being unrefreshed during the day. In recent years, there has been a significant increase in the occurrence of insomnia accompanied by pain and exhaustion. It is one of the most prevalent ailments in urban societies, yet patients and doctors often lack knowledge about it. There are no widely accepted standard treatments for insomnia. The DSM-5 criteria for insomnia (Mansukhani *et al*., 2014) include dissatisfaction with sleep quantity or quality, difficulty initiating sleep (in children, this may manifest as difficulty initiating sleep without caregiver intervention), difficulty maintaining sleep, characterized by frequent awakenings or difficulty returning to sleep after waking, and daytime impairment due to sleepiness. Although some restless nights are usual for most people in response to stress or noise, insomnia is often a secondary symptom of other medical conditions rather than a standalone disorder. It can lead to daytime tiredness, irritability, an increased risk of workplace accidents, difficulty operating machinery, and concentration lapses while driving.

One of the most crucial elements for identifying abnormalities in clinical EEGs and comprehending functioning activities in cognitive research is frequency. Human EEG potentials are aperiodic, unpredictable oscillations with intermittent bursts of oscillations that typically occur in several specifically pre-defined bands, such as 0.5-4 Hz (delta), 4-8 Hz (theta), 8-13 Hz (alpha), 13-30 Hz (beta), and >30 Hz (gamma) (Elgendi *et al*., 2011). The sources of these oscillations are billions of examples of these EEG cycles, as shown in Figure 2 (Büyükgöze, 2019).

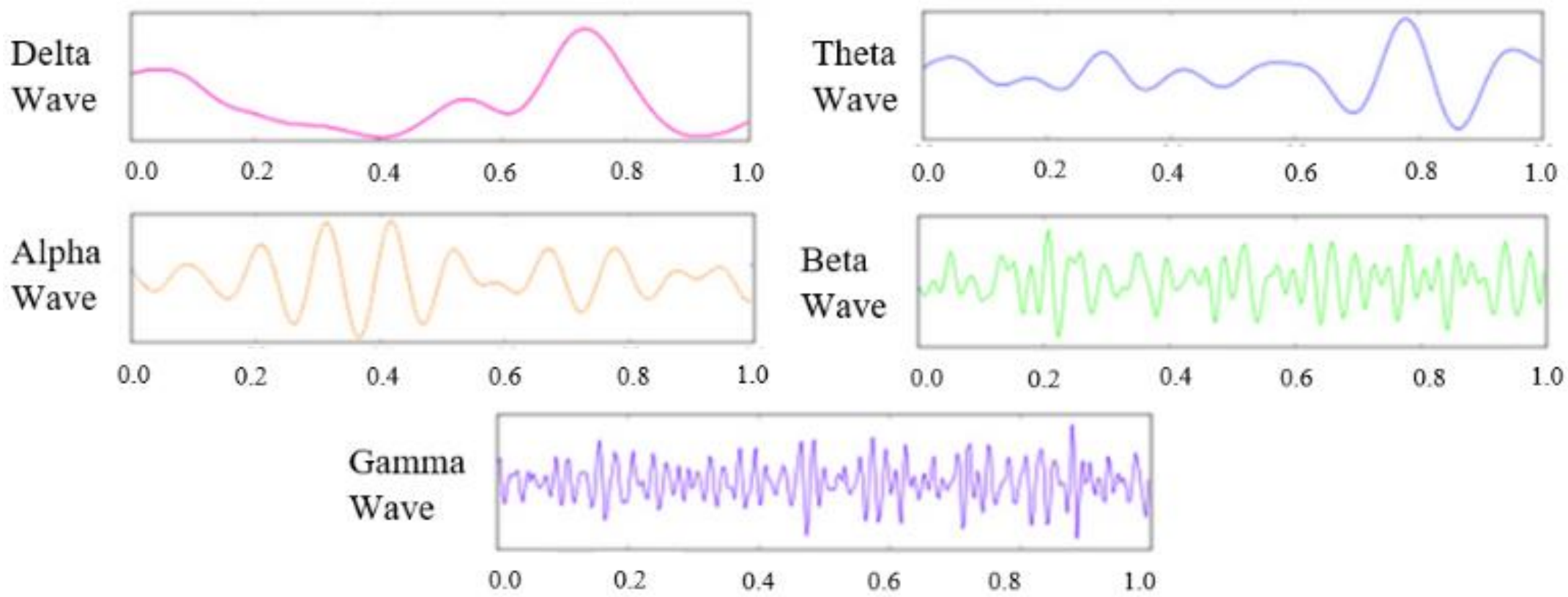

Figure 2 Examples of EEG rhythms - from top to bottom: delta, theta, alpha, beta and gramma rhythms (Büyükgöze, 2019).

Modern machine learning techniques such as convolutional neural networks (CNNs) are helpful for biomedical engineering applications, especially with regard to bio-signals, as they yield results that are easy to understand. However, most programming libraries, publications, and research are designed primarily for image-based applications, leading to the misconception that CNNs are unsuitable for non-image-based applications and hindering the broader adoption of CNNs in biomedical engineering. Although Electroencephalography (EEG) signals are inherently one-dimensional, and learned filters are commonly used to interpret these signals, there still needs to be a consensus on the best techniques among the signal processing and machine learning communities. In recent years, there has been a growing interest in applying CNNs to one-dimensional bio-signals (Kiranyaz *et al*., 2016; Cecotti *et al*., 2014).

**2. Literature Review**

Most current research on insomnia compares the characteristics of the EEG channel for insomnia patients to those for healthy individuals. Most features of these studies are linked to sleep depth and EEG spectral characteristics (Wu *et al*., 2013). When sleep is prominent, the EEG frequencies are highly dependent on the depth of sleep, as shown in (Hord *et al.*, 1965). High-frequency bands (*e.g.*, β and γ bands) that associate with increased cortical activity occasionally exhibit in wake, stage 1, and REM sleep. In a previous study, Hord *et al*. (1965) showed that insomnia can be observed using EEG spectral power in the β range, a cortical hyperarousal index. Buysse (2008) and Perlis *et al*. (2001) showed that patients with insomnia have a higher beta power than healthy individuals.

In a former study, Perlis *et al*. (2001) showed that insomnia causes an increase in the EEG σ and β power during NREM sleep. Wu *et al.* (2013) found no substantial differences in the sigma and beta strength between groups with insomnia and healthy groups during the wake and NREM periods and that insomnia patients have more inconsistent NREM and REM sleep than healthy subjects. Compared to healthy subjects, the patients with insomnia have disturbed sleep and substantially less slow-wave sleep (SWS) and REM sleep. In order to compare patients with insomnia to those who sleep well, researchers from Svetnik *et al.* (2017) used a vast database of sleep (NREM) electroencephalogram recordings and a power spectrum analysis while concurrently considering these variables and their interactions. Comparisons of power were made between cohorts (patients with insomnia versus good sleepers) by sex, considering age and the time of night. Researchers discovered that for both males and females, absolute power in the delta, theta, and sigma bands decreased with age. Women were inherently more powerful than men at all ages for each band, cohort, and phase of the night. These gender inequalities were substantially more extensive than those between insomniacs and sound sleepers. Perlis *et al.* (2021) analyzed the high-frequency EEG activity (HFA) in primary and secondary insomnia from a temporal and stage-wise perspective. Primary insomnia (PI), insomnia after major depressive disorder (MDD), and good sleeper controls (GS) were the three groups that were compared. Age, sex, and body mass were matched between the groups. For each sleep epoch, average spectral profiles were produced. Following the elimination of waking, moving, and arousal epochs and epochs containing micro or mini arousals, grand averages were calculated for each NREM cycle and each stage of sleep. It was discovered that HFA tends to rise throughout NREM cycles, and peaks happen during stage 1 and REM sleep. Both the effects are accentuated in patients with PI. Furthermore, HFA was discovered to be negatively correlated with Delta activity, and the three groups in our investigation seemed to display typical Delta/Beta patterns.

The statistical features of EEG readings reveal significant differences between insomnia diagnoses. Hamida *et al*. (2015a) used a *k*-means-based algorithm to detect insomnia automatically by introducing the higher-order Hjorth statistical parameters (Hjorth, 1970) as features for input variables. Most insomnia studies concentrate on characteristics of the spectral domain (Parrino, 2004; Neu *et al*., 2015).

Two studies by Wu *et al*. (2013) and Buysse (2008) examined the statistical features of EEG signals to determine differences between healthy individuals and those with insomnia. Spectral analysis (Buysse, 2008) of sleep EEG is not suited to the non-stationary nature of an EEG, but the wavelets are suitable (Hazarika *et al.*, 1997); however, the high computational overhead with the massive datasets is required. Since the wavelets can extract knowledge only in the time and frequency domains, the Hijorth parameters (Hjorth, 1970) were introduced to overcome the weaknesses and improve the accuracy of sleep stage classification (Vidaurre *et al*., 2009). Hamida *et al*. (2015a) used only Hjorth parameters with C3 and C4 channels for insomnia identification. According to the study, the C3 channel can reliably detect insomnia epochs from patients' records with a Cohen's kappa coefficient (Cohen, 1968) of 0.83, a sensitivity of 91.9%, and a specificity of 91.0%, so the study only used time domain features to classify insomnia patients and healthy subjects.

Frequency domain-based features are often used to classify sleep stages. However, there is a high computational complexity when using this approach for real-time applications. This study seeks to use both time domain and frequency domain features for sleep stage classification to reduce computational and time complexity,

A support vector machine (SVM) classifier was employed by Mulaffer *et al*. (2017) to develop two distinct models that distinguish between healthy people and insomnia patients. Data from 124 subjects, 70 of whom were healthy and 54 of whom had insomnia, were used to validate the models. The first model, which draws 57 features from two channels of EEG data (including statistical measures, Hjorth parameters, amplitude measures, and spectral features), has an accuracy of 81%. The second model's accuracy, which uses 15 characteristics from each participant's hypnogram and selects the best features based on F-value, is 74%. The methods for that study and this present study both use spectral and temporal domain features, but this study does not use sleep stage annotation data. Based on a dual-modal physiological feature fusion, Zhang *et al*. (2019) developed a system to automatically identify people with slight trouble falling asleep. The plan employed the random forest (RF) algorithm for classification and the correlation-based feature selection (CFS) method for gathering the ideal feature subset.

Shahin *et al*. (2017) employed a set of 57 EEG parameters (time-frequency domain features) generated from a maximum of two EEG channels in their study. They adopted deep learning to identify insomnia patients from healthy volunteers. Two different strategies were used to accomplish the goal. The second method used only EEG data from sleep phases impacted by insomnia, whereas the first method used EEG data for the entire sleep session, regardless of sleep state. Forty-one healthy participants and foty-one patients with insomnia participated in the study to evaluate the gadget. Compared to a manual assessment, the discrimination accuracies of both the groups utilizing two (C3 and C4) channels and one (C3) channel reached 92% and 86%, respectively. The authors of this study introduced sleep staging information for classifying insomnia.

Aydın *et al*. (2009) applied Artificial Neural Network (ANN) with spectra from Singular Spectrum Analysis (SSA) to diagnose insomnia using both C3 and C4 sleep EEG. The EEG signals, having different singular spectra with different sleep stages, were evaluated ahead with a sleep stager. Hamida *et al*. (2015b) developed an insomnia detection algorithm based merely on deep sleep stage signals acquired from the EEG C3 channel. To specify the signals only in the deep sleep stage, a sleep staging procedure was thus required. Latterly, based on single-channel labeled EEG signals with this kind of sleep stage annotation procedure, Yang and Liu (2020) proposed a one-dimensional convolutional neural network (1D-CNN) model to identify insomnia automatically. Qu *et al*. (2021) used alternatively a transfer learning for the classification when data scarcity happened to the detection. The study trained a sleep stage classifier to extract generally related features for modeling the classification. By confirming with the studies, a sleep staging step ahead of the major function of insomnia identification was conventionally required. Due to the importance of the sleep staging classifier, Zhao *et al*. (2022) evaluated algorithms developed based on single-channel EEG.

One challenge of this study is to classify patients as healthy or suffering from insomnia using only raw features derived from the time and spectral domains rather than relying on traditional methods.

## 3. Methodology

### 3.1. System description

This study aims to detect insomnia in a home setting using single-channel EEG monitoring. While single-channel EEG is often used for automatic sleep staging, its use in automatically identifying insomnia is rare. Sleep is a vital physiological function that impacts both physical and mental health (Ohayon, 2002), and although the AASM has established criteria for identification (Edinger *et al*., 2004; Newell *et al*., 2012), insomnia is still often diagnosed through imprecise and subjective questionnaire methods or time-consuming and expensive overnight PSG monitoring that cannot be performed at home (Fallmann and Chen, 2019). A single-channel EEG insomnia identifier could therefore be a key component in building a practical, home-based identification system and provide insight into the underlying causes of insomnia. Hardware for home-based sleep monitoring can include smart mats with piezoelectric and pressure sensors, ECG and pulse waves, or EEG (Fallmann and Chen, 2019), and various algorithms have been developed to detect sleep disorders automatically based on this hardware. As EEG signals are the gold standard for analyzing sleep activity and fewer channels in an EEG signal are the most promising way to identify sleep disorders, this study proposes using a 1D-CNN model with a high discrimination potential to detect insomnia automatically using fewer EEG channels. The flowchart in Figure 3 shows the individual modules of this study.

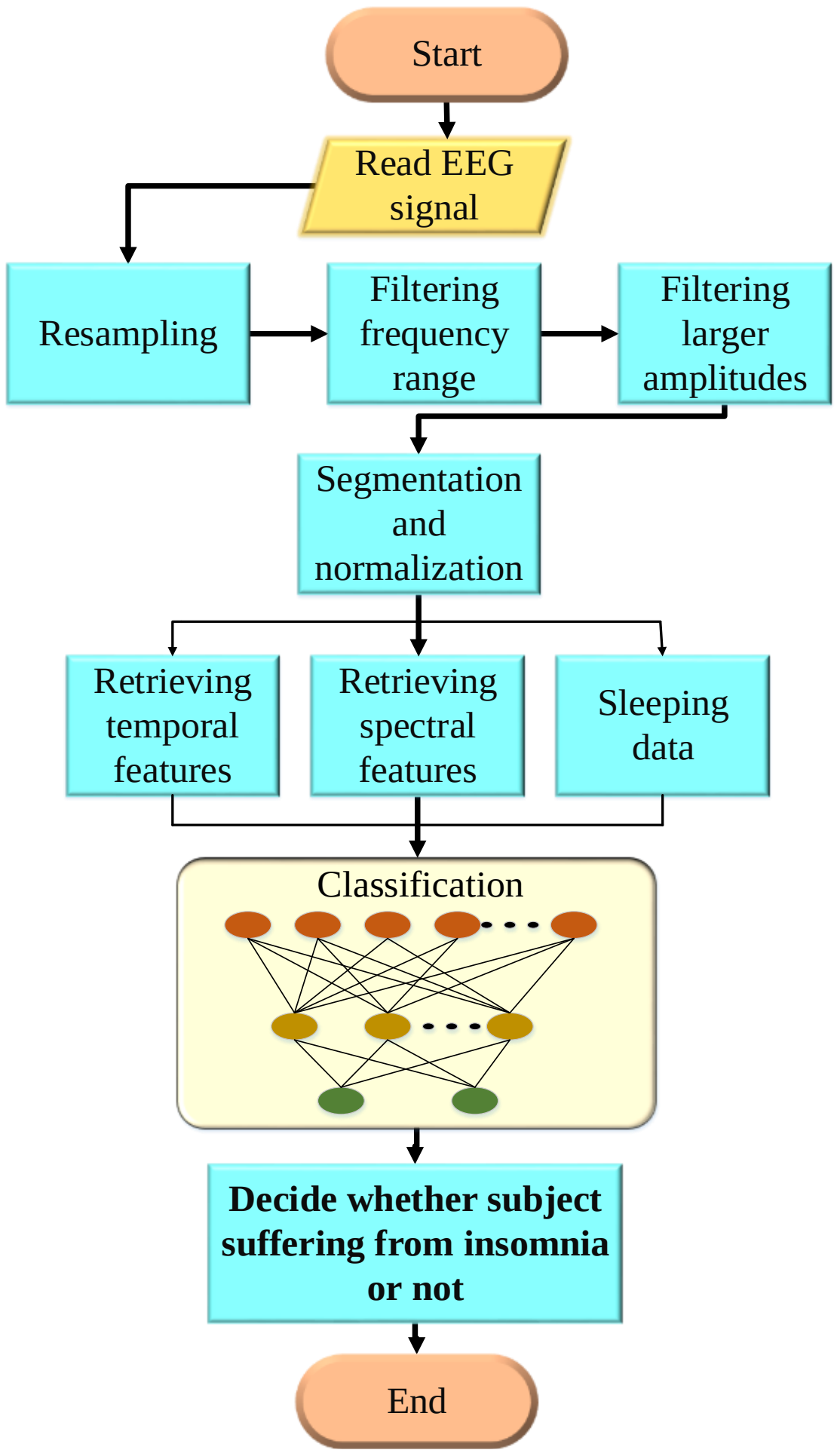

Figure 3  Flowchart for the system to automatically identify insomnia symptoms.

### 3.2. Preprocessing

The initial selection of C4 and Fp2 EEG channels for candidate signal acquisition is supported by medical reasoning. During sleep, brain activity typically originates in the occipital lobe, and moves toward the frontal lobe, so activity in the parietal and frontal lobes of the brain is more closely associated with sleep and deep sleep stages. Insomnia patients typically have less deep sleep than healthy subjects, so C4 and Fp2 channels are chosen for analysis to investigate insomnia.

To remove artifacts, the EEG data was filtered to 0.5-40 Hz using a 7th-order Butterworth filter (see Figure 4). Epochs with signals of a greater amplitude (> 260 μV) are eliminated using a straightforward out-of-bound voltage clipper because the main culprits of these abnormalities in the EEG data are muscle and/or ocular movements (Mansukhani *et al*., 2014). The EEG signal is then normalized to remove any DC shift that might have happened while the data was being recorded.

The preprocessing involves mainly filtering frequency ranges, filtering amplitudes, and normalization. The steps of preprocessing and their background motivations are detailed in Figure 4.

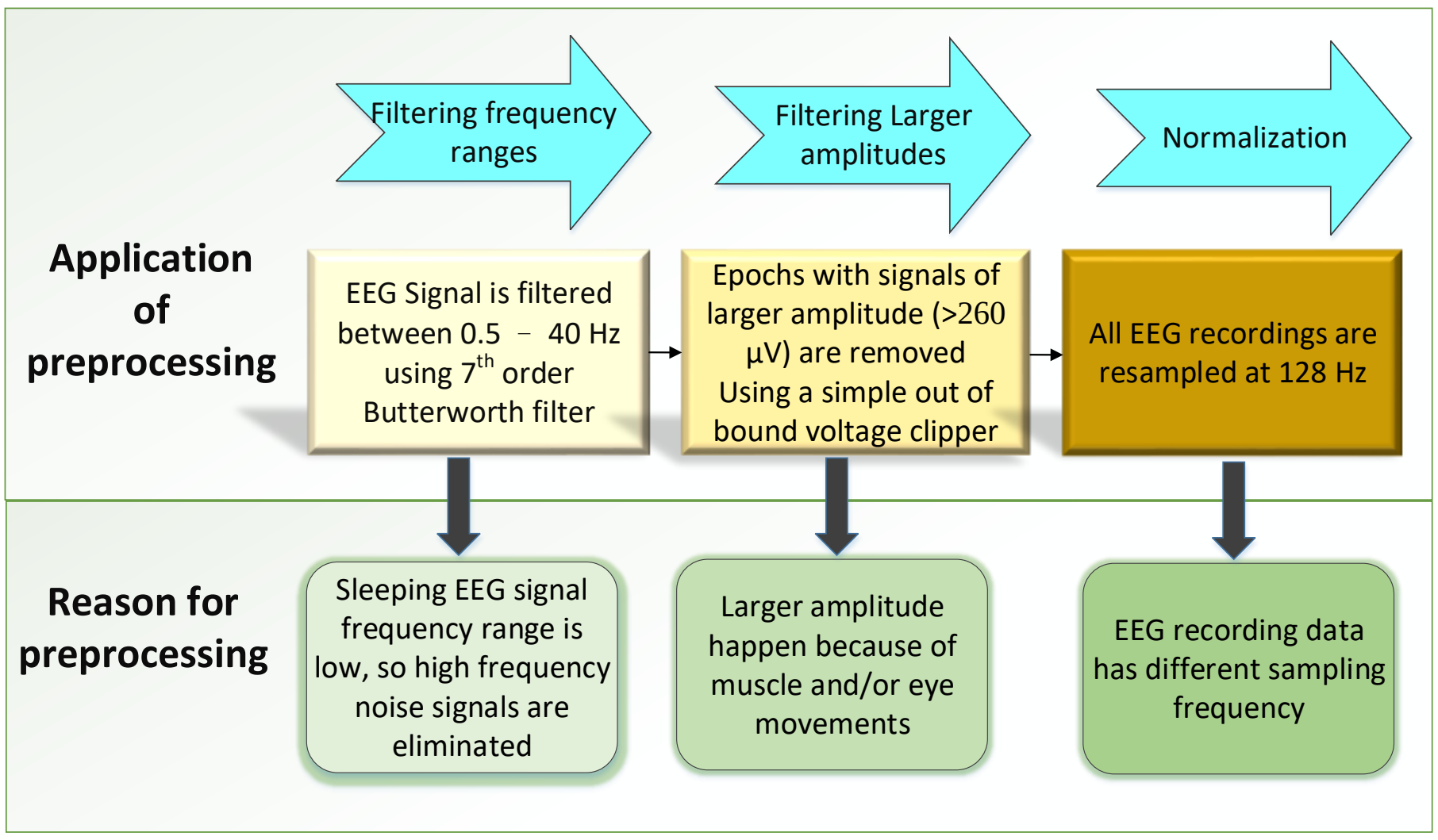

Figure 4 EEG signal preprocessing step.

As considered, the study implementations are intended to be applicable even if the data is sampled with different sampling rates, as the examples illustrated in Table 1. A frequency-consistent resampling procedure is thus introduced accordingly. To manifest a relatively high resolution for signals, a 128 Hz resampling frequency is adopted in the study. As the most significant common factor, the adopted resampling frequency renders a sufficient resolution of signals for further feature extraction.

Table 1 Example EEG recordings of sleep data.

| Subject code | Sleep duration | | Sampling frequency |
|---|---|---|---|
| | Start time | End time | (Hz) |
| Healthy-1 (CAP) | 22:19:06 | 06:38:36 | 512 |
| Healthy-2 (CAP) | 23:06:12 | 07:26:12 | 512 |
| Healthy-3 (CAP) | 22:36:37 | 07:01:37 | 100 |
| Healthy-4 (CAP) | 22:49:48 | 07:13:18 | 512 |
| Healthy-5 (TMU) | 22:12:22 | 06:15:02 | 1450 |
| Healthy-6 (TMU) | 22:10:24 | 05:48:14 | 1450 |
| Healthy-7 (TMU) | 22:21:07 | 06:10:12 | 1450 |
| Insomnia-1 (CAP) | 22:15:32 | 06:07:58 | 256 |
| Insomnia-2 (CAP) | 22:43:04 | 05:42:34 | 512 |
| Insomnia-3 (CAP) | 22:28:44 | 07:13:14 | 200 |
| Insomnia-4 (TMU) | 22:05:45 | 05:55:14 | 1450 |
| Insomnia-5 (TMU) | 22:09:33 | 05:42:33 | 1450 |
| Insomnia-6 (TMU) | 22:15:28 | 06:02:14 | 1450 |

### 3.3. Feature extraction

Information conveyed by the EEG signals is then summarized into a reduced set of measurements and is used to identify and recognize the target. As collectively surveyed, the features can be deduced majorly into three categories, based on their source characteristics, namely: (1) temporal (time) domain-based features, (2) spectral (frequency) domain-based features, and (3) non-linear features. Several effective time domain and frequency domain features, as listed in Table 2, are evaluated for this study. A zero mean and unit standard deviation normalization is applied to all the features to reject measurement variability among the subjects. With evaluation, 30 candidate features, which are selected from different domains: 24 from the spectral domain and 6 from the temporal domain, are admissible to represent the problem.

Table 2 List of extracted features.

| Source domain | Feature |
|---|---|
| Temporal domain | − Mean value (*MEAN*)<br>− Standard deviation (*STD*)<br>− Zero-crossing rate (*ZCR*)<br>− Hjorth parameters (*Activity*, *Mobility*, and *Complexity*) |
| Spectral domain | − Total power<br>− Absolute slow wave power<br>− Relative power in frequency bands of δ, θ, α, σ, β and γ<br>− Absolute power in frequency bands of δ, θ, α, β and γ<br>− Ratios of: (δ/θ), (δ/α), (δ/γ), (δ/β), (θ/α), (θ/γ), (θ/β), (α/γ), (α/β) and (γ/β) |

**Mean value (MEAN) and standard deviation (STD):** The descriptive statistics *MEAN*, and *STD* abbreviated for every designated 30s epoch are introduced:

$$MEAN = \frac{1}{N}\sum_{i=1}^{N} x_i \tag{1}$$

$$STD = \sqrt{\frac{1}{N}\sum_{i=i}^{N}(x_i - MEAN)^2} \tag{2}$$

where $N$ is the number of data points in the EEG signal and $x_i$ represents the amplitude of the EEG signal at the $i^{th}$ data point.

**Zero crossing rate (ZCR)**: As a signal with a high zero-crossing rate is frequently related to a sleep spindle, *ZCR* is adapted to locate and identify potentially the N2 sleep stage:

$$ZCR = \frac{1}{2(N-1)}\sum_{i=1}^{N-1} |\mathrm{sgn}(x_i) - \mathrm{sgn}(x_{i+1})| \tag{3}$$

where sgn($x$) denotes the sign function in which −1 is returned if $x$ is negative, 0 if $x$ is zero, and 1 if $x$ is positive.

**Hjorth parameters**: As ever been elevated in insomnia patients, the Hjorth parameters (Hjorth, 1970), including *Activity*, *Mobility*, and *Complexity*, are time domain statistical properties, which are calculated from the variance of a signal or the first-order derivative of the signal, are alternatively indicators to represent the EEG amplitude, gradient and gradient spread, respectively:

$$Activity = var(\mathbf{x}) \tag{4}$$

$$Mobility = \frac{Activity\,(\frac{\Delta \mathrm{x}}{\Delta t})}{Activity\,(\mathbf{x})} \tag{5}$$

$$Complexity = \frac{Mobility\,(\frac{\Delta \mathrm{x}}{\Delta t})}{Mobility\,(\mathbf{x})} \tag{6}$$

where **x** represents the signal train [$x_0$, $x_1$, …, $x_{N-1}$] of the designated epoch. While the *Activity* represents the signal power and the *Mobility* represents an estimate of the mean frequency, the *Complexity*, indicating the similarity of the appearance of the signal to an exact sine wave (Vidaurre *et al.*, 2009), represents the change in frequency.

**Spectral domain features**: Power spectral density (PSD), a DTFT of an autocorrelation sequence, is a typical spectral domain signal processing approach that displays the strength of the energy as a function of frequency and distributes the signal power over frequency. The Welch method (Ng *et al*., 2019), a modified segmentation strategy, is employed to assess the spectral power in the study. For the $N$ point data sequence in the designated epoch, $x(n)$, $n$ = 0, 2, …, $N$–1, a modified periodogram is:

$$P_{\mathbf{x}}(f) = \frac{1}{NU}\left|\sum_{n=0}^{N-1} x(n)w(n)e^{-j2\pi nf}\right|^2 \tag{7}$$

where $w(n)$ is the window function, $U = 1/N\sum_{n=0}^{N-1} w^2(n)$ is a normalization modifier for the periodogram calculation. The PSD of the EEG signal at frequency $f$ is given by averaging the periodograms over the spectral resolution of the epoch, which is of length $L$:

$$PSD_{\mathbf{x}}(f) = \frac{1}{L}\sum_{i=0}^{L-1} P_x(f) \tag{8}$$

Then, the absolute power of each frequency band is determined by:

$$Absolute_Power_{\mathbf{x}b} = \int_{f_{\mathrm{b,min}}}^{f_{\mathrm{b,max}}} PSD_{\mathbf{x}}(f)df \tag{9}$$

where [$f_{\mathrm{b,min}}$ $f_{\mathrm{b,max}}$] specifies the band b's frequency range. Alternatively, the total power can also be calculated by (9) while [$f_{\mathrm{b,min}}$ $f_{\mathrm{b,max}}$] specifies the entire frequency range (0.5 – 45 Hz) of the sampled signals. Hereafter, the relative power, adopted to represent the normalized spectral power concerning variations among various sleep stages, is defined as a ratio of the absolute power to the total power:

$$Relative_Power_{\mathbf{x}b} = Absolute_Power_{\mathbf{x}b}/Total_Power \tag{10}$$

Moreover, the power ratio of $b_1/b_2$ is calculated in the same way by changing the division base instead.

**Absolute slow wave power**: The slow wave, typically ranging from 0.5 Hz to 2.0 Hz with a signal amplitude greater than 75 μV, often regards deep sleep in the subjects. The power of it, calculated by (9), is thus also recruited for evaluation.

### 3.4. Classification

The proposed CNN architecture uses relatively large kernel sizes of 1×3 and 1×2 for the first and second convolution layers. Small kernels with a size of 1×1 are consecutively deployed in the third and fourth convolution layers to minimize the size of feature maps. The feature maps generated for the final maximal pooling layer are conventionally flattened into a column vector. The final identification result is produced by feeding this vector into the fully connected layer for binary classification. Table 3 lists the relevant hyperparameters for the CNN, including the layer type, the number of kernels, the size of adopted kernels, the convolution stride for each convolutional layer, the size of the pooling region, the convolution stride for each pooling layer, and the number of units for each fully connected layer.

Table 3 Parameters of adopted CNN model.

| Layer ID | Layer type | No. eliminate kernel | Kernel size | Region size | Stride | Padding | Output size |
|---|---|---|---|---|---|---|---|
| 1 | 1D-Conv | 32 | 1x2 | - | 1 | - | 32x1x18 |
| 2 | 1D-Conv | 32 | 1x2 | - | 1 | - | 32x1x17 |
| 3 | MaxPool | 32 | - | - | - | no | 32x1x8 |
| 4 | 1D-Conv | 128 | 1x1 | - | 1 | - | 128x1x8 |
| 5 | MaxPool | 128 | - | - | - | no | 128x1x4 |
| 6 | 1D-Conv | 256 | 1x1 | - | 1 | - | 256x1x4 |
| 7 | MaxPool | 256 | - | - | - | no | 256x1x2 |
| 8 | Flatten | - | - | - | - | - | 512x1 |
| 9 | Dense | - | - | 512 | - | - | 1x512 |
| 10 | Dense | - | - | 128 | - | - | 1x128 |
| 11 | Dense | - | - | 2 | - | - | 1x2 |

## 4. Experiment data

### 4.1. Cyclic Alternating Pattern (CAP) sleep dataset

The CAP Sleep database (Terzano *et al.*, 2001; Goldberger *et al.*, 2012) provided part of the data for this study. PSG recordings for 108 individuals were collected in the database, including healthy persons, people with insomnia, and people with various sleep disorders. Only the healthy (25 records) and insomnia (9 records) subject's data were extracted for this study. Expert neurologists who were well-trained at sleep centers provided the scoring of the sleep macrostructure according to the R & K rules into the six stages of Sleep (Wake, S1-S4, REM, and Body movement).

### 4.2. Taipei Medical University hospital dataset

This study's second batch of data is obtained from the sleeping center at Taipei Medical University (TMU) Hospital in Taiwan. All the patients' data in this part are primarily collected under the patient agreements (Sleep Clinic TMU Hospital, 2021). They were transferred after de-identification under permission for the secondary analysis. The dataset comprises PSG recordings for sleep-disorder patients which includes 2 EOG channels, 1 EMG for the submentalis muscle, 1 respiration signal, 1 ECG signal, and 4 EEG channels with a 30s epoch for the recordings.

### 4.3. Sample characteristics

These two datasets are used to construct the study dataset, which includes 8-hour overnight recordings for 50 healthy subjects and 50 insomnia cases. There are 46 females (27 healthy subjects and 19 insomnia patients) and 54 males (23 healthy subjects and 31 insomnia patients) with a median age of 46 years, ranging from 37 to 67. This study uses EEG channels C4 and Fp2 with a sampling frequency greater than 200 Hz to extract the feature values. Table 4 lists the sample characteristics with the mean and deviation to describe their corresponding distribution.

Table 4 Summary of sample characteristics.

| Characteristic | All patients (*n*=100) | Healthy (*n*=50) | Insomnia (*n*=50) |
|---|---|---|---|
| Age (years) | 51.7 ± 13.8 | 52.4 ± 13.1 | 49.6 ± 10.8 |
| Sex (male) | 54% | 42.60% | 56.40% |
| Weight (lbs.) | 194.5 ± 51.7 | 207.8 ± 54.8 | 176.2 ± 51.7 |
| BMI (kg/m$^2$) | 27.4 ± 8.8 | 21.2 ± 2.6 | 30.6 ± 5.8 |
| Total sleep time (min) | 429 ± 123 | 510 ± 102 | 349 ± 144 |
| Sleep efficiency (%) | 79.8 ± 18.5 | 85.4 ± 12.5 | 64.2 ± 24.6 |

## 5. Experimental Results and Discussion

### 5.1. Signal preprocessing

To show the first denoising process of the signals, Figures 5 and 6 depict EEG signals from the C4 and Fp2 channels with noise (before filtering) and without noise (after filtering), respectively. The signals in Figures 5(a) and 6(a) have higher variance (amplitude range > 256 µV). This variance is indeed from the contaminated noise. The signals in Figures 5(b) and 6(b) contrast the noise-removed clearer signals having lower variance.

Individual δ, θ, α, β, and γ bands of components are extracted from the denoised signals for feature selection. Figure 7 shows the frequency bands that are extracted from the denoised signal of Figure 6(b).

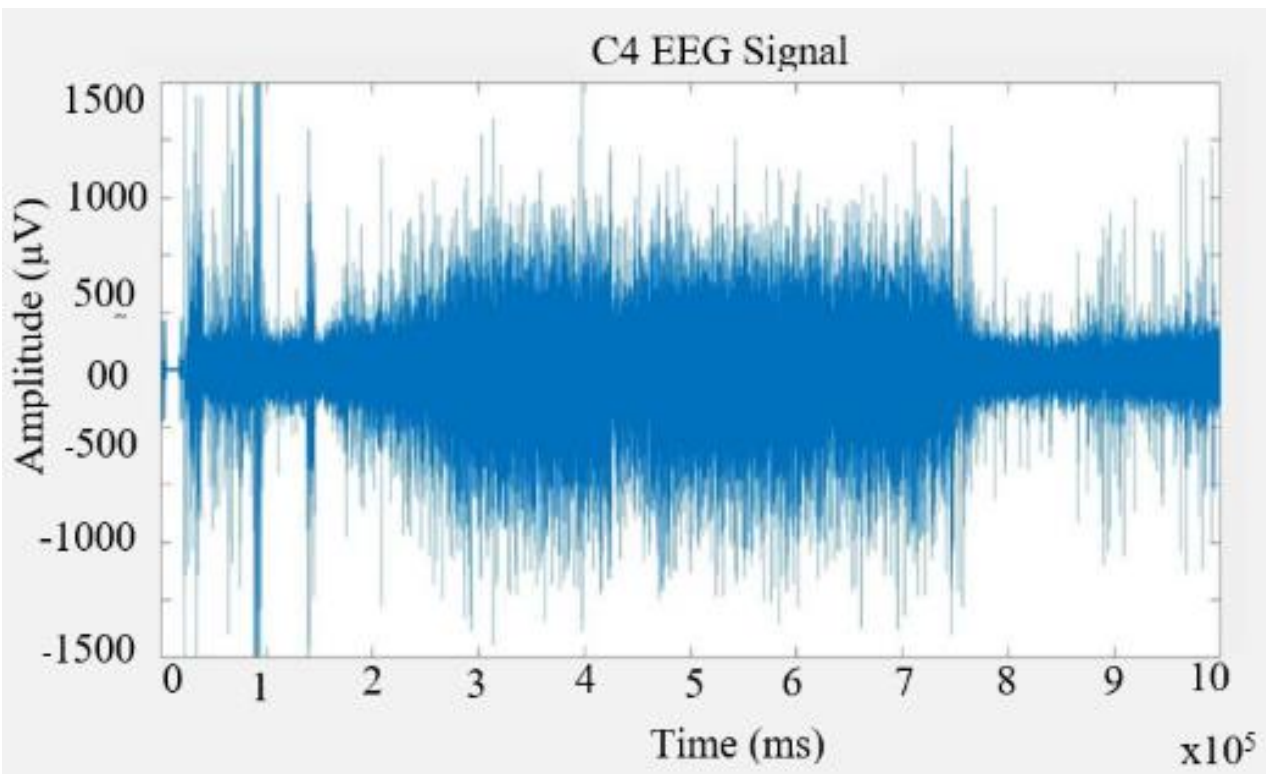

(a) Before denoising

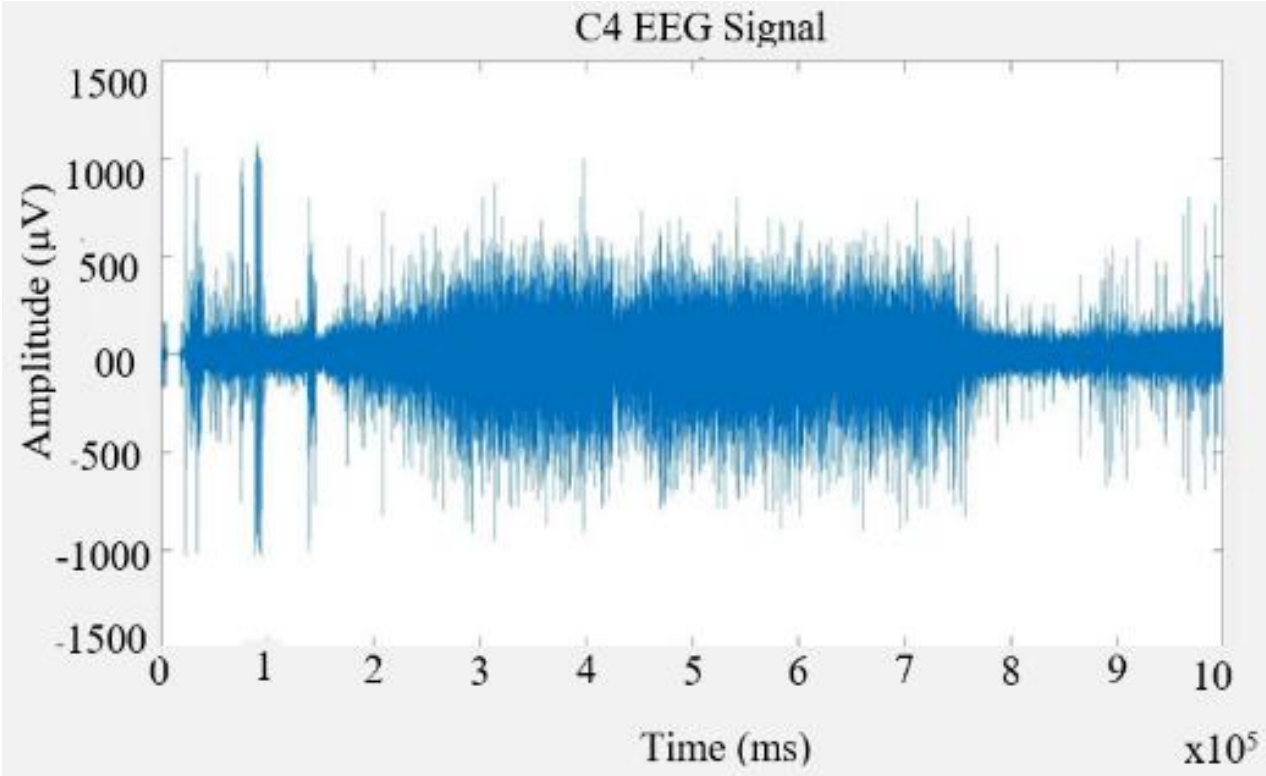

(b) After denoising

Figure 5  Sample EEG signal from C4 channel.

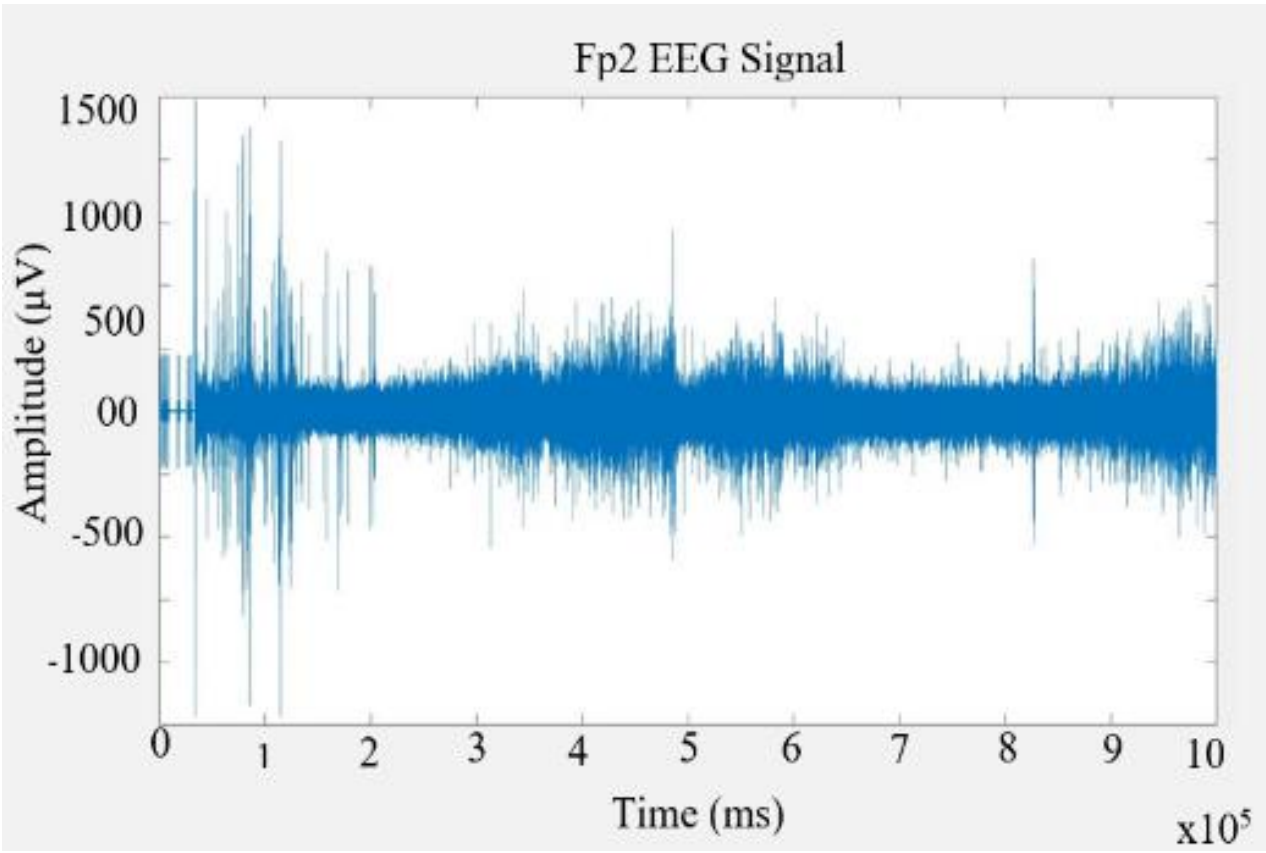

(a) Before denoising

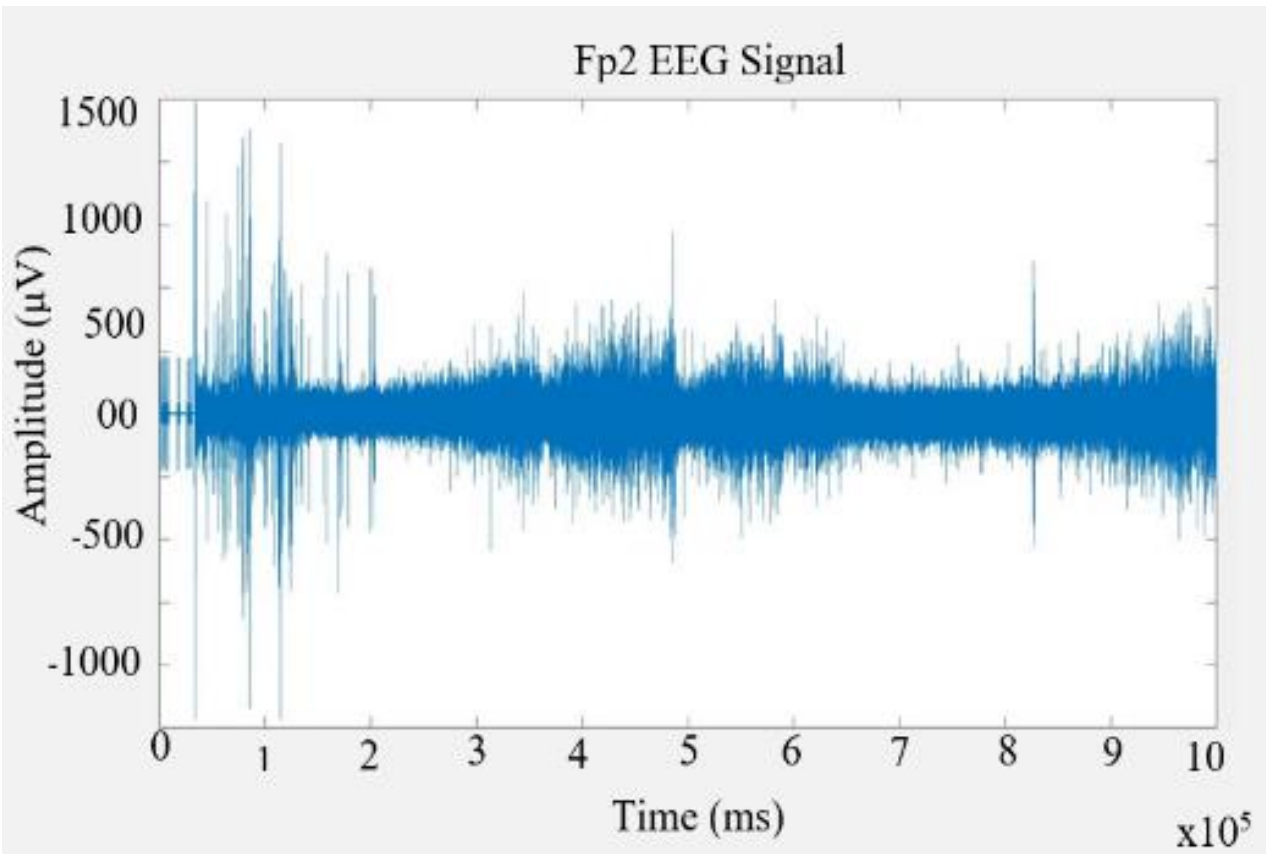

(b) After denoising

Figure 6  Sample EEG signals for the Fp2 channel.

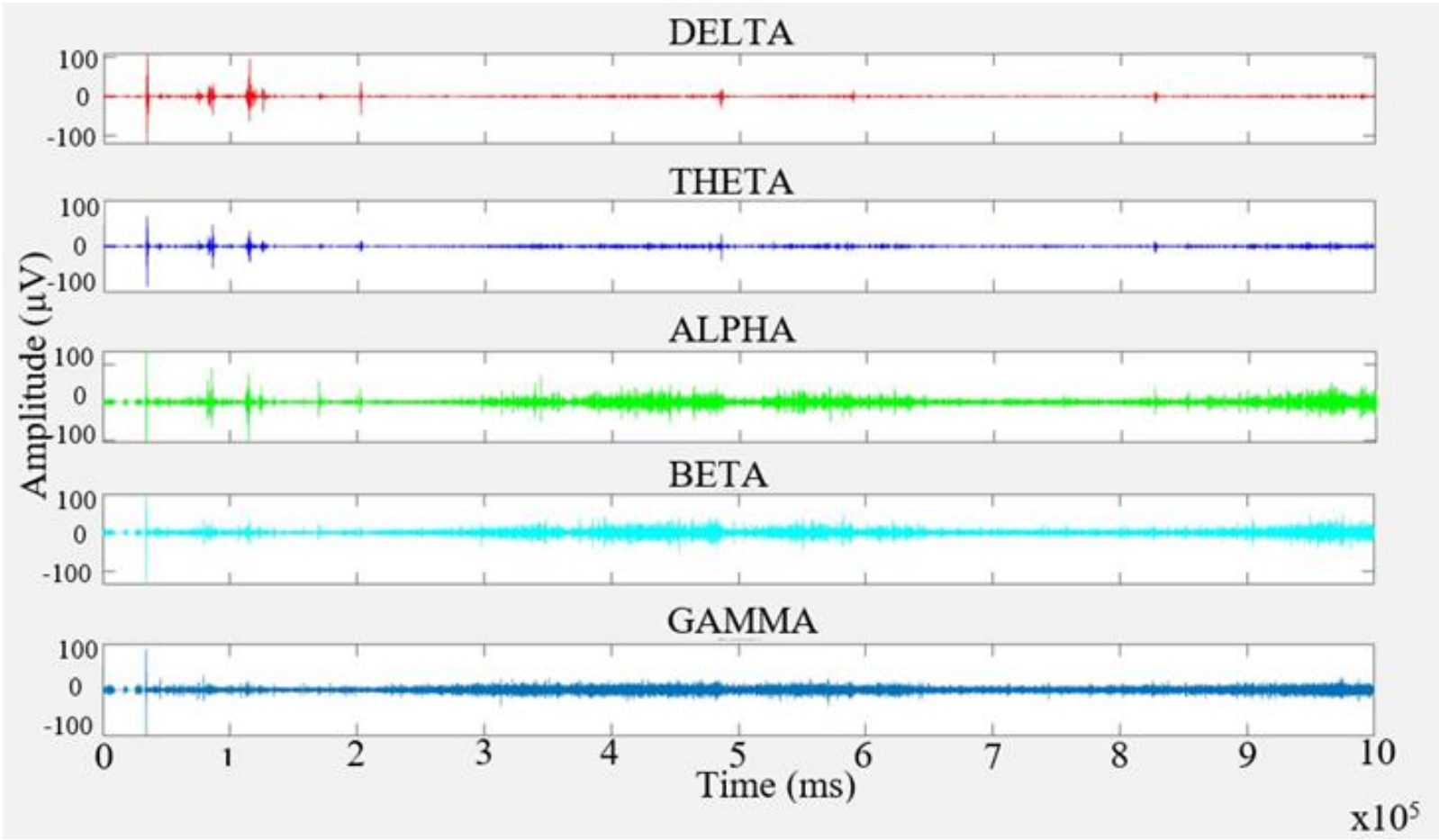

Figure 7  EEG signal segmented into $\delta$, θ, α, β and γ components for a specific frequency domain.

### 5.2. Feature selection

A process for selecting pseudo-optimal features is introduced to determine the optimal feature set that accurately quantifies the number of EEG features for classification. Exhaustively searching all possible combinations to choose the best feature set would be computationally expensive and time-consuming, so a statistics-based feature selection method is used to quantitatively and qualitatively evaluate and prioritize candidate features. This process selects a subset of features that best differentiate between the goals and does not require all possible combinations of candidate attributes. The priority determines a pseudo-optimal combination of features that allows for effective and efficient classification. As validated, the recommended pseudo-optimal feature set is effective in accurately classifying insomnia patients from healthy subjects.

*1) Pearson correlation*

A Pearson correlation analysis determines the correlation between the target categories, Insomnia and Healthy, and each candidate feature is employed here to evaluate the relevant priority. The Pearson correlation coefficient measures the degree to which the relationship between the paired variables. The coefficient would vary within the range [–1, +1]. While 0 implies no correlation between the variables, a coefficient closer to –1 or +1 implies an extremely strong linear relationship. The resultant coefficient values for features to pair with the category tag (*CAT*, valued either Insomnia or Healthy) are listed in Table 6, where the 30 candidate features and their annotations are listed in Table 5.

Table 5 Candidate features' notation and annotation.

| Feature division | Notation | Annotation |
|---|---|---|
| Relative power | $P_{r\gamma}$ | Relative γ power |
| | $P_{r\beta}$ | Relative β power |
| | $P_{r\sigma}$ | Relative σ power |
| | $P_{r\alpha}$ | Relative α power |
| | $P_{r\theta}$ | Relative θ power |
| | $P_{r\delta}$ | Relative δ power |
| Absolute power | $P_{asw}$ | Absolute slow wave power |
| | *TP* | Total power |
| | $P_{a\gamma}$ | Absolute γ power |
| | $P_{a\beta}$ | Absolute β power |
| | $P_{a\sigma}$ | Absolute σ power |
| | $P_{a\alpha}$ | Absolute α power |
| | $P_{a\theta}$ | Absolute θ power |
| | $P_{a\delta}$ | Absolute δ power |
| Ratio power | $PR_{\gamma/\beta}$ | Ratio power γ/β |
| | $PR_{\alpha/\beta}$ | Ratio power α/β |
| | $PR_{\alpha/\gamma}$ | Ratio power α/γ |
| | $PR_{\theta/\beta}$ | Ratio power θ/β |
| | $PR_{\theta/\gamma}$ | Ratio power θ/γ |
| | $PR_{\theta/\alpha}$ | Ratio power θ/α |
| | $PR_{\delta/\beta}$ | Ratio power δ/β |
| | $PR_{\delta/\gamma}$ | Ratio power δ/γ |
| | $PR_{\delta/\alpha}$ | Ratio power δ/α |
| | $PR_{\delta/\theta}$ | Ratio power δ/θ |
| Time domain | *ZCR* | Zero crossing rate |
| | *MEAN* | Mean value |
| | *STD* | Standard derivation |
| | *Activity* | Defined in (1) |
| | *Mobility* | Defined in (2) |
| | *Complexity* | Defined in (3) |

Table 6 Pearson correlation coefficients between category tag (*CAT*) and the candidate features.

| Correlation between *CAT* and Relative power | | Correlation between *CAT* and Absolute power | | Correlation between *CAT* and Ratio power | | Correlation between *CAT* and Time domain parameter | |
|---|---|---|---|---|---|---|---|
| | *CAT* | | *CAT* | | *CAT* | | *CAT* |
| $P_{r\gamma}$ | .61 | $P_{asw}$ | .71 | $PR_{\gamma/\beta}$ | .43 | *ZCR* | .68 |
| $P_{r\beta}$ | .54 | *TP* | .25 | $PR_{\alpha/\beta}$ | .26 | *MEAN* | .1 |
| $P_{r\sigma}$ | .03 | $P_{a\gamma}$ | .33 | $PR_{\alpha/\gamma}$ | .30 | *STD* | .23 |
| $P_{r\alpha}$ | .57 | $P_{a\beta}$ | .28 | $PR_{\theta/\beta}$ | .33 | *Activity* | .26 |
| $P_{r\theta}$ | .41 | $P_{a\alpha}$ | .28 | $PR_{\theta/\gamma}$ | .36 | *Mobility* | .41 |
| $P_{r\delta}$ | .64 | $P_{a\theta}$ | .26 | $PR_{\theta/\alpha}$ | .11 | *Complexity* | .41 |
| *CAT* | 1 | $P_{a\sigma}$ | .21 | $PR_{\delta/\beta}$ | .41 | *CAT* | 1 |
| | | $P_{a\delta}$ | .15 | $PR_{\delta/\gamma}$ | .31 | | |
| | | *CAT* | 1 | $PR_{\delta/\alpha}$ | .67 | | |
| | | | | $PR_{\delta/\theta}$ | .50 | | |
| | | | | *CAT* | 1 | | |

### *2) T-test and p value*

A t-test compares two averages and determines the significance level if they are different, determining whether these differences are due to chance. This study uses a paired sample t-test to determine the features differentiating insomnia patients from healthy subjects. A *p*-value is a number that describes how likely the probability of the null hypothesis will be true. The level of statistical significance for a t-test is often expressed by a real valued probability *p*. The smaller the probability *p*, the stronger the evidence is to accept the null hypothesis. The calculated t statistics and *p*-values for each candidate feature with *CAT* are shown in Table 7. The critical t values are also shown to determine the feature relevance.

Table 7 The t-value and *p*-value for all features.

| Feature | t-statistic | Critical t-value | *p*-value | Feature | t-statistic | Critical t-value | *p*-value |
|---|---|---|---|---|---|---|---|
| $P_{r\gamma}$ | 1.27 | 1.23 | 0.24 | $PR_{\delta/\gamma}$ | 1.26 | 1.20 | 0.25 |
| $P_{r\beta}$ | 0.33 | 0.33 | 0.75 | $PR_{\delta/\beta}$ | 1.54 | 1.46 | 0.16 |
| $P_{r\sigma}$ | 0.17 | 0.33 | 0.90 | $PR_{\theta/\alpha}$ | 0.02 | 0.02 | 0.98 |
| $P_{r\alpha}$ | 0.44 | 0.42 | 0.68 | $PR_{\theta/\gamma}$ | 1.56 | 1.47 | 0.16 |
| $P_{r\theta}$ | **2.44** | **2.38** | **0.03** | $PR_{\theta/\beta}$ | 1.64 | 1.56 | 0.14 |
| $P_{r\delta}$ | 1.55 | 1.50 | 0.15 | $PR_{\alpha/\gamma}$ | 1.25 | 1.18 | 0.25 |
| $P_{a\delta}$ | 0.74 | 0.75 | 0.47 | $PR_{\alpha/\beta}$ | 1.36 | 1.29 | 0.21 |
| $P_{a\theta}$ | 0.83 | 0.84 | 0.41 | $PR_{\gamma/\beta}$ | 1.79 | 1.79 | 0.10 |
| $P_{a\alpha}$ | 0.93 | 0.93 | 0.36 | *MEAN* | 0.42 | 0.41 | 0.69 |
| $P_{a\beta}$ | 1.00 | 1.00 | 0.33 | *STD* | 0.65 | 0.65 | 0.53 |
| $P_{a\gamma}$ | 1.17 | 1.17 | 0.26 | *ZCR* | **3.28** | **3.22** | **0.01** |
| *TP* | 0.84 | 0.85 | 0.41 | *Activity* | 0.95 | 0.95 | 0.35 |
| $P_{asw}$ | **2.90** | **2.88** | **0.01** | *Mobility* | 0.41 | 0.39 | 0.40 |
| $PR_{\delta/\theta}$ | 0.68 | 0.65 | 0.52 | *Complexity* | 0.52 | 0.50 | 0.63 |
| $PR_{\delta/\alpha}$ | 0.36 | 0.36 | 0.72 | | | | |

### *3) Optimal feature set selected*

All these candidate features are used to test the null hypothesis for the data of healthy subjects and insomnia patients. Table 7 shows the relevance of the priority for feature selection. By evaluating the tabulated data in Table 7, 18 features of the 30 features are selected, and two rules are established to rate the features:

[1] *If* (the t statistic is significantly greater than the critical t value) *and* (the probability *p* is less than 0.05) and (the Pearson correlation is significantly high with *CAT*) *then* (select as a top feature)

[2] *If* (the t statistic is greater than the critical t value) and (the probability $p$ is less than 0.70) *and* (the Pearson correlation is high with *CAT*) *then* (select as an optimal feature)

Highly relevant features are then identified, including the relative power of the $\delta$, θ, β, and γ bands, the total power, the absolute slow wave power, the power ratios $\delta$/θ, $\delta$/α, $\delta$/γ, $\delta$/β, θ/α, θ/β, α/γ and α/β, the mean, the zero crossing rate, the mobility, and the complexity, which amounts to 18 features. The top three optimal features that differentiate insomnia patients from healthy subjects are the zero crossing rate *ZCR*, the absolute slow wave power $P_{asw}$ and relative θ power $P_{r\theta}$, highlighted in bold font in Table 7.

### 5.3. Distinguishability with direct features

The relevant features and stage parameters for healthy subjects and insomnia patients are examined using descriptive statistics (mean, standard deviation, minimum, and maximum values) to check for distinguishability. The results in bold font in Tables 8 and 9 highlight those significantly distinguishable. The highlighted values in the tables demonstrate that the relevant features are comparable to the stage parameters in distinguishability, confirming that sleep staging information can be excluded from the identification process.

Table 8 Mean and standard deviation of relevant features and stage parameters.

| Sleep parameters | Healthy subjects | | Insomnia patients | |
|---|---|---|---|---|
| | Mean | Std | Mean | Std |
| ZCR (count) | **26,468.96** | 38,308.62 | **161,167.3** | 155,625.3 |
| $P_{asw}$ (μV) | **401,529.96** | 712268.1 | **279567.28** | 382025.5 |
| $P_{r\theta}$ (ratio) | **$5.58\times10^{-1}$** | $1.64\times10^{-1}$ | **$3.50\times10^{-1}$** | $1.84\times10^{-1}$ |
| Stage wake (s) | **2,798.75** | 1,904.57 | **13,856.67** | 9,436.21 |
| Stage S1 (s) | 969.31 | 1,133.31 | 1,491.33 | 1,431.87 |
| Stage S2 (s) | 13,800.37 | 2,609.14 | 12,344.33 | 4,454.36 |
| Stage S3 (s) | 3,118.37 | 1,368.75 | 3,361.33 | 1,610.63 |
| Stage S4 (s) | **5,177.00** | 2,210.78 | **3,052.50** | 1,052.55 |
| Stage REM (s) | **6,854.75** | 1,656.15 | **4,158.66** | 19,260.55 |

Table 9 Maximum and minimum of relevant features and stage parameters.

| Sleep parameters | Healthy subjects | | Insomnia patients | |
|---|---|---|---|---|
| | Min | Max | Min | Max |
| ZCR (count) | **1,456** | **271,069** | **15031** | **824,418** |
| $P_{asw}$ (μV) | **58,274.66** | **844,995.7** | **59,631** | **2161,732** |
| $P_{r\theta}$ (ratio) | **$3.79\times10^{-1}$** | **$8.80\times10^{-1}$** | **$1.51\times10^{-1}$** | **$7.46\times10^{-1}$** |
| Stage wake (s) | **300** | **6,690** | **1,800** | **25,100** |
| Stage S1 (s) | 60 | 7,070 | 1,500 | 4,379 |
| Stage S2 (s) | 8,690 | 17,205 | 6,607 | 17,900 |
| Stage S3 (s) | 1,450 | 17,012 | 1,737 | 6,748 |
| Stage S4 (s) | **2,611** | **9,967** | **1,316** | **4,154** |
| Stage REM (s) | **8,910** | **1,140** | **1,200** | **6,990** |

Figure 8 shows the data from Tables 8 and 9 in boxplots. The panels demonstrate significant differences in features between healthy subjects and insomnia patients. All the findings in Figure 8 are consistent with initial expectations and confirm the feasibility of direct identification without sleep staging information.

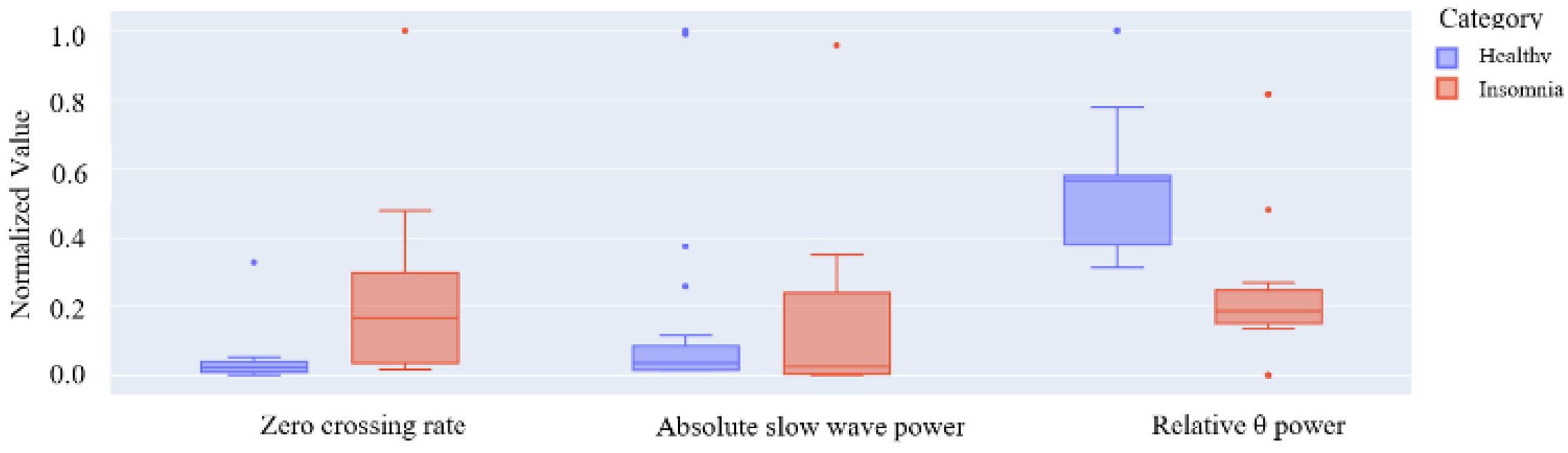

Figure 8 Boxplots of the relevant features in Tables 8 and 9: healthy subjects represented in blue color and insomnia patients in red.

### 5.4. Performance tuning and evaluation

*1) Performance metrics*

The measurement of the efficacy of identification is not only accuracy but also the statistics of precision, recall, F1 score, and Cohen's kappa coefficient ($k$). While the famous metrics of accuracy, precision, recall, and F1 score are listed in (4)-(7), respectively (Fawcett, 2006), the Cohen's kappa coefficient measures the degree of agreement between two referees who separately assign things to mutually exclusive categories is defined in (8) (Cohen, 1968). In the expressions, $TP$, $TN$, $FP$, and $FN$ denote the count number of true positives, true negatives, false positives, and false negatives in the confusion matrix.

$$Accuracy = \frac{TP+TN}{TP+TN+FP+FN} \tag{4}$$

$$Precision = \frac{TP}{TP+FP} \tag{5}$$

$$Recall = \frac{TP}{TP+FN} \tag{6}$$

$$F1\ score = \frac{2 \times Precision \times Recall}{Precision + Recall} \tag{7}$$

$$k = \frac{2(TP \cdot TN - FN \cdot FP)}{(TP+FP) \cdot (FP+TN) + (TP+FN) \cdot (FN+TN)} \tag{8}$$

*2) Classifier selection*

The system for this study uses a Jupiter notebook environment with Python programming language. For both Fp2 and C4 channel datasets, 75% of the selected feature data record is randomly chosen as the training set, and the left 25% is the test set. The CNN model is fitted for the training and test data using batch size 1. The training and validation accuracies are then plotted for various iterations for performance evaluation.

With a hyperparameter tuning procedure, the Fp2 channel data achieve the best training accuracy of 99.78% and validation accuracy of 97.85% with a learning rate = $3 \times 10^{-4}$, a weight decay = 0, and an Adam optimizer. The C4 channel data, with a learning rate = $2 \times 10^{-4}$ and a weight decay = 0, and an Adam optimizer achieve a training accuracy of 99.52% and validation accuracy of 93.84%. The training and validation curves for both datasets obtained using the CNN model are shown in Figure 9. The training and validation losses gradually decrease and converge to a stable state within precise ranges when the learning process matures.

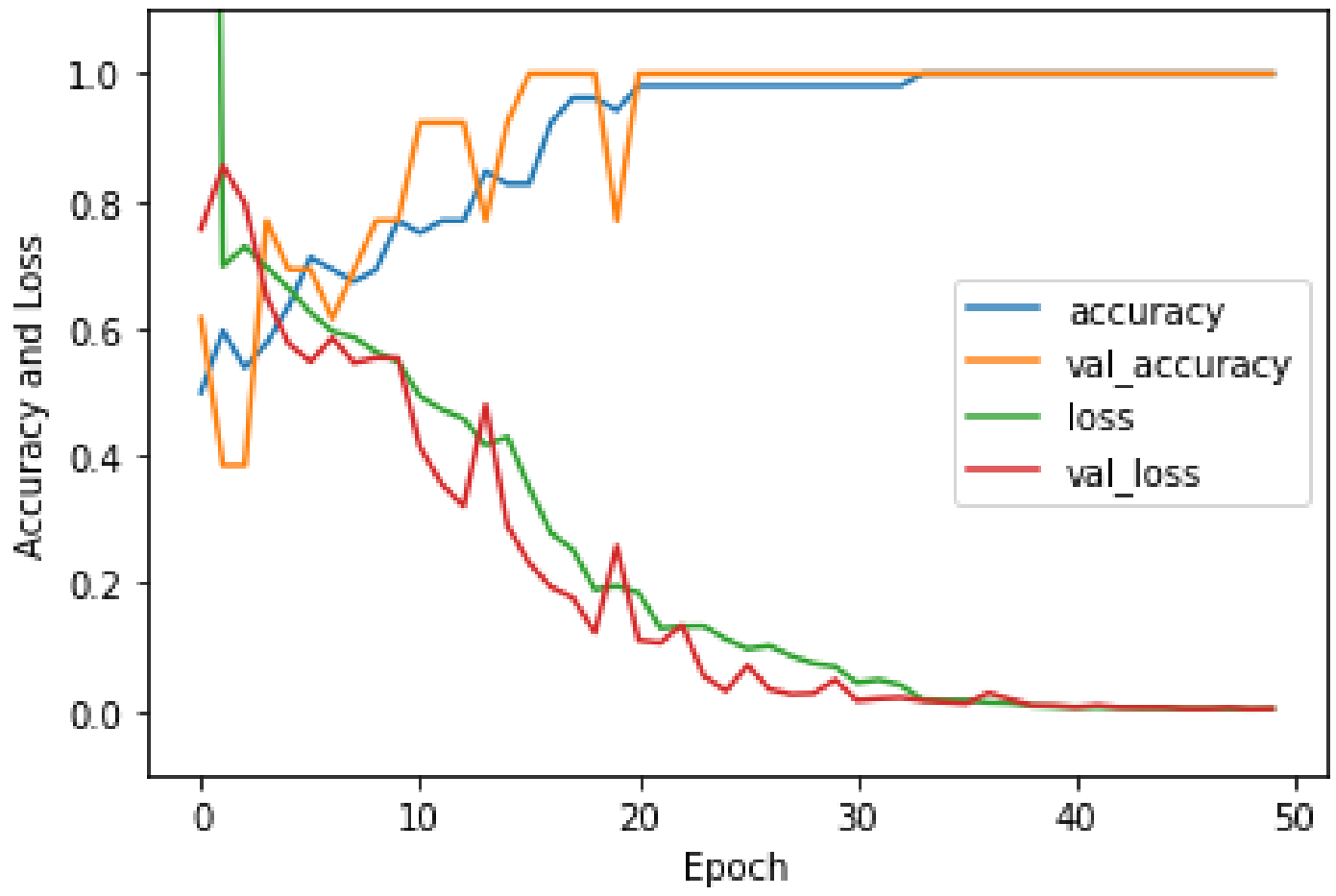

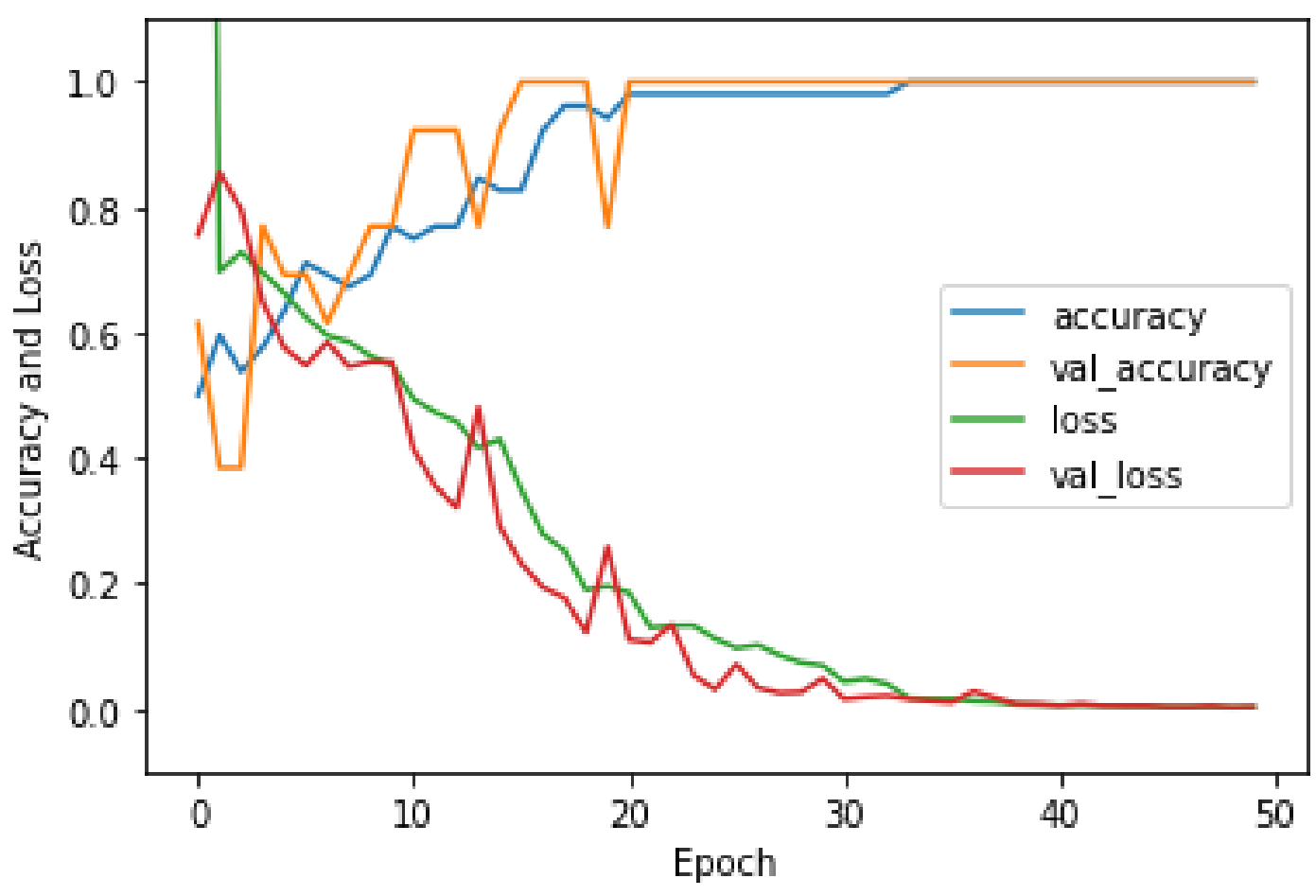

Figure 9 The accuracy of the Fp2 (upper panel) and C4 (lower panel) channels and the loss at each epoch during the learning process.

Identification using the optimal features from the Fp2 channel and the best-tuned CNN model is 97.85 % accurate, has 96.14% precision, 100% recall, 96.25% F1 score, and 94.15% Cohen's kappa value. A comparison with the results for other classification algorithms, including a random forest (RF), an ANN, and a support vector machine (SVM), is shown in Table 10. The results show that the CNN for this study outperforms other baseline models.

Table 10 Performance (%) comparison between different classifiers.

| Method | Accuracy | Precision | Recall | F1 score | *k* |
|---|---|---|---|---|---|
| CNN | **97.85** | 96.14 | 100 | 96.25 | 94.15 |
| RF | 94.12 | 100 | 88.47 | 94.28 | 87.35 |
| ANN | 95.78 | 100 | 88.53 | 95.78 | 89.85 |
| SVM | 92.53 | 94.35 | 94.57 | 93..69 | 93.72 |

*3) Anticipation of optional probe combination*

The EEG channel combination that best identifies insomnia is then determined. The data for EEG channels Fp2 and C4 can be used, so using the data for both channels may increase performance. An experiment compares the results for different combinations of channels, *i.e.*, C4 only, Fp2 only, and C4+Fp2. The classification results of the optional combinations are shown in Table 11.

Table 11 Performance (%) evaluations for classifiers with different combinations of channels.

| Channel / Evaluation parameters | Fp2 | C4 | C4 + Fp2 |
|---|---|---|---|
| Accuracy | **97.85** | 93.84 | 97.65 |
| Precision | 96.14 | 87.28 | 95.75 |
| Recall | 100 | 100 | 100 |
| F1 score | 96.25 | 94.85 | 97.85 |
| *k* | 94.15 | 88.78 | 94.76 |

Table 11 shows that Fp2 only performs better than the other two options: C4 and C4+Fp2. During sleep, slow wave activities (frequency 0.5-2 Hz and min amplitude 75 μV) and delta wave activities (frequency 1-4 Hz) mostly happen in the frontal area. Slow waves and delta waves are associated with the deep sleep stages, stage 3 (S3 and S4), and REM. In stage 3, less than half of the brain's activity is composed of delta waves, whereas during REM sleep, more than half of the brain's activity is dominated by delta waves. The results in Table 11 show a significant difference in the duration of REM and S4 for healthy subjects and insomnia patients. The frontal EEG channel better differentiates healthy subjects from insomnia patients, so the features that are extracted from the EEG frontal electrodes allow a substantial degree of discrimination.

*4) Performance comparison with other studies*

The results for an optimally tuned version of the proposed insomnia identification scheme are compared with those for other studies, as shown in Table 12. As compared, the analysis methods of other studies used their corresponding sleep stage annotation. The additional sleep stage annotation makes the identification more time-consuming. For real-time classification, a method with less complexity and high accuracy will benefit more. Table 12 reveals that the proposed model enables classification without a sleep stager and outperforms existing methods.

Table 12 Performance of different identification methods.

| | Classifier | Accuracy (%) | *k* (%) | Characteristic |
|---|---|---|---|---|
| Shahin *et al.* (2017) | DNN | 86 | 71 | With sleep stage annotation |
| Aydin *et al*. (2009) | ANN | 76.85 | 52 | |
| Hamida *et al.* (2016) | PCA | 78.57 | 55 | |
| Yang and Liu (2020) | CNN | 91.31 | 82 | |
| Sharma *et al*. (2021) | EBDT | 89.56 | - | |
| Zhang *et al.* (2019) | RF | 96.22 | 62 | Without sleep stage annotation |
| The proposed method | CNN | **97.85** | 94.15 | |

## 5. Conclusion

This study introduced a classification model to automatically identify insomnia using single-channel EEG data without sleep stage annotation. The proposed model was competitively efficient compared to surveyed models, not only for the reduced procedure complexity of no staging annotation but also for improved identification performance. Data from C4 and Fp2 channels and their combination were evaluated, and single-channel EEG at Fp2 was found to have the highest performance. The study also explained signal preprocessing, feature extraction, feature selection, and classification procedures. Using overnight PSG recording data for evaluation, the validated identification performance reached an accuracy of 97.85% and a Cohen's kappa coefficient of 94.15. The valid performance of the proposed model paved the way for developing a home-use insomnia identification device using a single-channel EEG.

**Acknowledgment**

The authors sincerely thank Professor Shih-Chin Chen and the Sleep Clinic at Taipei Medical University Hospital for their help in using the data for modeling and validation.